# Optimizing Local Explanations in Private Tree Ensembles

Sonal Allana[1], Sam Fletcher[2], Rozita Dara[1], and Pulei Xiong[3]

[1]University of Guelph, Canada

[2]Litera, Canada

[3]National Research Council of Canada

Corresponding author: Rozita Dara (email: drozita@uoguelph.ca).

This research was funded by a Natural Sciences and Engineering Research Council of Canada (NSERC) Discovery Grant awarded to Dr. Rozita Dara. It is partially supported by collaborative research funding from the National Research Council of Canada's Artificial Intelligence for Logistics Program.

**ABSTRACT**

Explainability, privacy and accuracy are three essential principles for achieving trustworthiness in AI systems. However, there exists an inherent conflict in achieving these principles simultaneously. In this research, inspired by the mechanistic interpretability approach for neural networks, we propose an explainability method that leverages intrinsic model components to build local post-hoc explanations for points of interest. We demonstrate the approach on tree-based ensemble models by applying multi-objective optimization objectives on the constituent decision trees using a combination of decision tree and rule properties. We study the effect of applying differential privacy and its impact on all three principles: local explainability, mitigation of membership inference attacks and accuracy. When using strong privacy guarantees, membership inference success rates are held at random guess and the local explanations maintain consistency with their non-private versions when larger number of trees are contained in the ensemble. Our proposed explanation consistency metrics enable the quantification of the drift in explanations in the presence of privacy enhancing technologies and are applicable to local and global explanation methods that use explicitly scoped rules.

**INDEX TERMS** explainability, multi-objective optimization, differential privacy, membership inference, mechanistic interpretability, XAI.

## I. INTRODUCTION

Building trustworthy AI systems requires guiding principles to mitigate the risks of deploying these systems in real world applications. First, the AI must be accurate, otherwise it will lead to incorrect decisions and a loss of trust. If accuracy is achievable, two more principles become important for gaining people's trust: privacy ensures the protection of personal data of individuals, and explainability gives transparency into the decisions of complex black-box systems. Though achieving these two principles should result in systems that are explainable and privacy compliant, recent research has shown an inherent conflict between the two, where the inclusion of explainability can result in the leakage of privacy [1]. Similarly, other research has shown that it is not enough to measure utility as a proxy for explainability, and that explainability should be measured directly [2].

Models with Explainable AI (XAI) capabilities are found to demonstrate an extended attack surface and susceptibility to various black-box and white-box privacy attacks through explanations. Due to this potential of XAI in leaking the privacy of individuals, the use of privacy preservation methods, such as data anonymisation, differential privacy, cryptography, perturbation and federated learning [3], are explored with different XAI strategies. However, the integration of privatization techniques can also result in adversely affecting the utility of the system, thus requiring a delicate balance between explainability, privacy and utility [4]. This mutual interaction between these three properties is an important consideration when deploying XAI methods in production environments.

The use of differential privacy (DP) in training of machine learning (ML) models, enables the reporting of privacy guarantees and has emerged as a gold standard for privacy accounting [5]. Current research explores strategies for application of DP for privacy preservation in XAI pipelines [6]–[8], however its effect on explainability has been limitedly studied. Previous work [9] on DP training in deep learning has noted non-negligible impacts on feature-based explanations, but currently we have a narrow understanding of the consistency of explanations under DP conditions for other explanation types. There is also a lack of evaluation metrics for quantifying this consistency, or the absence of it, with different XAI techniques.

The explainability of models can be global or local in scope. While global explanations focus on the behaviour of the entire model, local explanations provide insights on individual inferences [10] and are useful when recipients of automated decisions seek the rationale for their respective outcomes. Thus, local explanations offer transparency and trust to the users of automated decision-making AI systems, and we focus on this aspect of Trustworthy AI in this work. These local explanations can be generated intrinsically from interpretable models or using post-hoc techniques from black-box models [11]. We envision a hybrid technique of explainability that combines both approaches to construct local post-hoc explanations using the intrinsic components of a model. We propose to apply model-specific optimization objectives to facilitate the selection of skillful internal components that can be consolidated to produce an explanation. To enable this screening of internal components, these optimization objectives must be based on influential factors, such as accuracy and complexity. We demonstrate this approach using classification tree-based ensembles based on majority voting and composed of individually interpretable trees, which, when combined together, render the model into a black-box.

Though current literature has proposed approaches for explainability of tree-based ensembles, to the best of our knowledge, none consider extraction of internal components using optimization of multiple model-specific objectives. This proposed optimization is aimed at selecting quality nodes and decision paths, yielding effective explanations and potentially improving their quality for end users. Further, we also study the effect of DP training of the underlying model on explanations and propose metrics to quantify their consistency under various privacy budgets.

In summary, the main contributions of this article are:

- Proposal of a local explainability approach using a selection of internal components based on model-specific optimization objectives in tree ensembles.
- Proposal of metrics for evaluation of consistency of explanations under DP.
- Explainability, privacy and utility analysis of the proposed algorithm under DP training.

The rest of this article is organized as follows. Section II provides the background context for the study, while Section III describes our proposed explanation method. In Section IV, we elaborate the experimental settings and evaluations on explainability, privacy and utility, while in Section V we discuss the results. We present related works in Section VI and conclude the article in Section VII.

## II. PRELIMINARIES

This section provides the background context of this research. We elaborate on privacy leakages in XAI-aware systems and the mathematical guarantees of DP. We explain our choice of a tree ensemble algorithm for the study and an applicable privatization algorithm. We also elaborate on the multi-objective optimization strategy applied in this work.

### A. PRIVACY LEAKAGES IN XAI

Though explainability is a fundamental principle of Trustworthy AI, recent research has found that explanations released through XAI interfaces are vulnerable to intentional and unintentional privacy leakages [3]. Researchers have demonstrated misuse of explanations for determining membership in the training set, exposure of sensitive features in the inputs and/or stealing of the model functionality.

Membership inference is a type of privacy attack in which model outputs are exploited to infer if an instance was part of the training set [12]. Model outputs, such as predictions and classifications, can be employed to execute this attack. When explanations are available, they are found to provide a strong attack surface and can be alternatively misused [1]. The attack is possible on different techniques of XAI and can lead to identification of individuals and exposure of their sensitive information. The attack is demonstrated on decision trees [13] in settings where they are used as surrogates with black-boxes. In such settings, the leakage through decision trees can exceed that of the corresponding black-boxes, with overfitted trees causing additional leakage.

### B. DIFFERENTIAL PRIVACY

DP provides a quantifiable definition of privacy and a means of tuning the privacy in a system. If $x$ is a sensitive dataset and $M(x)$ is the output generated by a property or mechanism $M$, and if $x$ is updated to $y$ by adding, updating or deleting a single record such that $M(x)$ is indistinguishable from $M(y)$, then $M$ is said to satisfy DP [14]. Mathematically, M gives ϵ-differential privacy if for all subsets S ∈ range(M) and ϵ > 0, the following definition is satisfied:

$$\frac{Pr(M(x) \in S)}{Pr(M(y) \in S)} <= e^{\epsilon} \qquad (1)$$

By adjusting the privacy budget, ϵ, from 0 to ∞, practitioners can manage the trade-off between privacy and utility [15]. DP possesses a post-processing property [16], according to which any data independent mapping of the mechanism is also differentially private with the same privacy budget, ϵ. Due to this property, the outputs of a DP trained model, including the intrinsically extracted explanations, also satisfy the privacy budget.

Since by design, the goal of DP is hiding the individual records in the dataset [17], it by definition limits the success rate of membership inference [18]. As a result, it is applicable for the empirical evaluation of the risk of membership inference of models trained under DP conditions.

## C. DECISION TREES AND TREE ENSEMBLES

Decision trees are hierarchical models that recursively split the input feature space into partitions based on feature values to classify samples [19]. Each internal node in the tree represents a condition on an input feature, and each leaf node represents a class label. During inferencing of a sample, the path from the root to a leaf node constitutes its explanation, and the class label at the leaf node is the predicted output. Tree ensembles are composed of multiple decision trees and combine the predictions from the constituent trees to improve accuracy and robustness [20]. Random forests, one such ensemble method, uses the class label predicted by majority of trees as the final output, also known as majority voting [21]. In these models, each tree is trained on different random samples of the training set [22] and has benefits such as resilience to overfitting and noise [21]. They have high predictive power [22] and are widely adopted [23] across different industries, such as healthcare, law, finance and security [24]. However, with increase in the number of trees in these models, the architectural complexity increases and deteriorates the overall interpretability of the models [20], thus driving the need for post-hoc approaches of generating explanations for the outcomes of random forest models.

## D. RANDOM FORESTS WITH DP

There are many ways to incorporate DP in decision tree based models [25]. Fletcher et al. [26] describe one such algorithm for building private random forests that optimizes for prediction accuracy. The privacy cost is minimized by randomly building the trees independently of any individual records, and only exposing the majority class label of each rule. This lets them calculate the majority vote across many trees without needing to pay the privacy cost for data-based queries at each node split, or for the support or confidence of leaf nodes. At each non-leaf node, the feature is selected randomly, and in the case of continuous features, so is the value to split the node on. While the trees remain interpretable, the authors claim the main downside of their method is that explainability is reduced, as the applicable rule in each tree for any given data point were arrived at coincidentally. As part of this work, we evaluate the explainability of random forests built using this method, and whether useful explanations can be extracted using our proposed method.

### E. FAITHFUL BY DESIGN APPROACH

Mechanistic interpretability seeks to reverse engineer models [27], specifically neural networks, by finding interpretable components that influence the outcomes. Our proposed explainability method is inspired by this approach and seeks to select such intrinsic interpretable units of a model post-hoc for specific instances. This generates local explanations that are faithful since the components are extracted from within the model architectures, rather than approximated. The challenge lies in identifying these model-specific interpretable units and aggregating the information from a multitude of such components for different architectures. For tree ensembles, we propose the use of decision nodes selected through an optimization of multiple tree-specific objectives. To determine simple yet skillful components, a combination of objectives that focus on performance, accuracy and complexity can be utilized.

### F. MULTI OBJECTIVE OPTIMIZATION

In a multi-objective optimization (MOO) problem, more than one objective are systematically and simultaneously optimized [28]. If **F(x)** is a vector of *k* objective functions, the MOO is given by:

$$\underset{x}{\text{Minimize}}\ \mathbf{F}(\mathbf{x}) = [f_1(x), f_2(x), ..., f_k(x)]^T \qquad (2)$$

subject to $g_j(x) \leq 0$ for $j = 1, 2, ..., m$, and $h_l(x) = 0$ for $l = 1, 2, ..., e$, where $m$ is the number of inequality constraints and $e$ is the number of equality constraints [28]. The solving of multiple conflicting objectives results in a large number of solutions, thus requiring a trade-off strategy to identify optimal solutions from a larger set of possibilities [29]. Pareto optimality is one such strategy that builds a set of points that are optimal, referred to as the Pareto front or Pareto frontier. A point, $x^*$, in the design space is called Pareto optimal, if and only if, there does not exist another point, $x$, such that $F(x) <= F(x^*)$ and $f_i(x) < f_i(x^*)$ for at least one $i$ [28].

The solutions in the Pareto frontier are non-dominated by each other [29], such that not one objective is improved without degrading another. Thus, this strategy provides a multi-dimensional multi-objective alternative to the single objective optimization problem [29].

## III. METHOD

In this section, we describe our model-specific local explainability approach for tree ensembles, including the choice of optimization objectives and the selection and aggregation of multiple interpretable components. We propose metrics for evaluation of explainability, privacy and utility when using DP at various privacy budgets. To aid understanding, we illustrate our approach with the help of an example.

## A. BACKGROUND

A rule is a path in a decision tree from the root to a leaf node and is of the format $\{C \Rightarrow T\}$[30] where, $C$ consists of a conjunction of conditions, given by $c_1 \cap c_2 \cap \ldots \cap c_n$, and $T$ is the outcome of the rule. Each condition, $c_i$, corresponds to a node in the rule and is of the format $f_i\ op_i\ t_i$, where $f_i$ refers to the $i^{th}$ input feature that influences an output. The operator, $op_i$, is either <= or >. The threshold, $t_i$, is the value of $f_i$ used to split the node. For example, *salary >20K*, is a condition on the input feature, *salary*, and indicates that its value should be greater than 20K for the condition to be true. We extract conditions from different trees within the ensemble and refer to the collection of selected conditions as a rulelist. These differ from rules since they are not actual paths from the root to leaf nodes and are composed of conditions extracted from multiple trees in the ensemble.

For the model-specific optimization objectives, the following properties of decision trees and corresponding rules are available:

- **Support:** This is the proportion of samples in the dataset that satisfy the antecedent of the rule. For the rule $\{C \Rightarrow T\}$, the support is calculated as follows,

$$\text{support} = \frac{\text{\#samples that satisfy C}}{\text{\#samples}} \qquad (3)$$

- **Confidence:** This is the proportion of samples in the dataset that are correctly classified by a rule out of the samples that satisfy the rule antecedent. It can be expressed as follows,

$$\text{confidence} = \frac{\text{\#samples that satisfy C} \Rightarrow \text{T}}{\text{\#samples that satisfy C}} \qquad (4)$$

- **Rule length:** This is the number of conditions contained in a rule. When $C$ consists of $n$ conditions, $c_1 \cap c_2 \cap \ldots \cap c_n$, the length of the rule is given by,

$$\text{rule length} = \text{n} \qquad (5)$$

- **Accuracy:** This is the ratio of the number of samples correctly classified by a decision tree to the total number of samples. This is a property of the decision tree rather than a rule and is given as follows,

$$\text{accuracy} = \frac{\text{\#correctly classified samples}}{\text{\#samples}} \qquad (6)$$

Support, confidence and accuracy can be computed on both data seen and not seen during training. Data not seen is often held out for testing, but also includes future real-world data. To differentiate between these computations, we refer to the latter as test support, confidence and accuracy. The DP algorithm for random forests by Fletcher et al. [26] avoids the privacy cost of data-based queries for support and confidence calculations. Hence to prevent incurring further privacy cost, publicly available test data or synthetically generated data can be used to compute test support and confidence instead [31]. In Section IV, we run an ablation study with both these types of test data to compare its corresponding effect on explanation quality metrics.

In accordance with the Minimum Description Length principle [32], we give preference to rules with lower complexity by minimizing the rule length. Both training and test sets are taken into consideration when setting the objectives of MOO, by measuring the accuracy of the trees on the training set while using the test set for support and confidence.

## B. PROPOSED EXPLAINABILITY ALGORITHM

Our proposed algorithm (Algorithm 1) comprises of three main steps: tree selection based on MOO, condition extraction, and rulelist compression. Figure 1 summarises the main flow of the algorithm and the steps are elaborated in the following subsections.

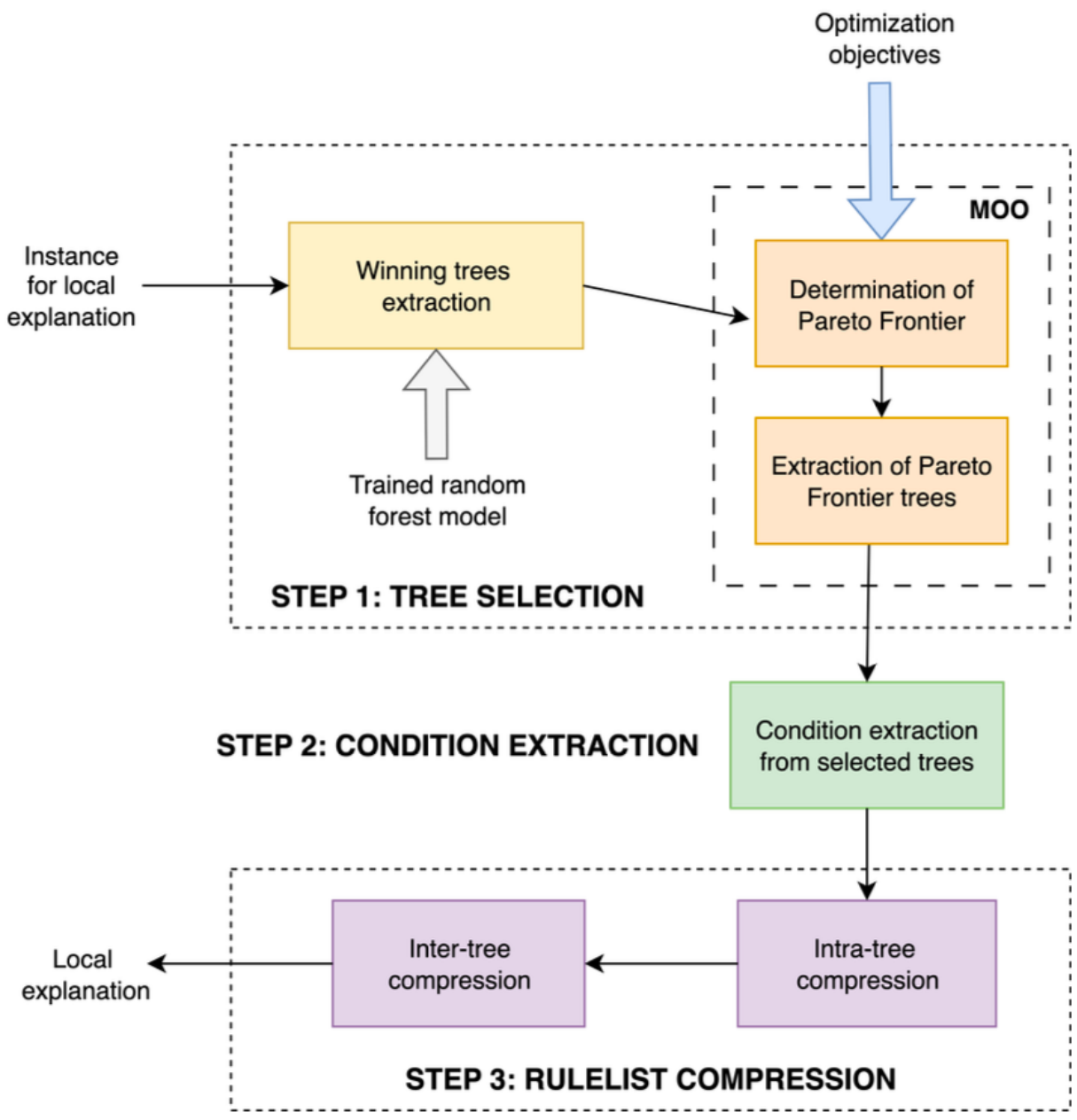

FIGURE 1: Local explanation generation.

### 1) TREE SELECTION

This step is used for the selection of a subset of trees from the ensemble. It enables the reduction of the complexity of the explanation model by screening trees that are suitable as the starting point for rule extraction. The selection is applied twice: once for exclusion of trees with outcomes not aligned with the ensemble and secondly, using MOO on a set of objectives. This two stage filtering enables the ranking of trees on select criteria while giving computational gains and complexity reduction.

We apply Pareto optimality [28] for MOO by selecting optimization objectives based on both decision tree and rule properties. This enables an implicit ranking and selection of good trees for rule extraction that balance multiple and conflicting objectives. The MOO estimates the corresponding Pareto frontier consisting of optimized trees based on the optimization direction of the respective objectives [33]:

$$\{\text{optimized trees}\} \underset{objectives}{\Longleftarrow} \{\text{winning trees}\} \qquad (7)$$

As identified in Section III A, among the rule properties, we select length of the rule (Eq. 5) and its support (Eq. 3) and confidence (Eq. 4) on the test set. Among the decision tree properties, we select accuracy on the training set (Eq. 6). To filter skillful trees, we maximize the training accuracy, test support and confidence. To minimize the number of conditions and hence the complexity, we minimize the length.

**Algorithm 1** Proposed explainability algorithm.

**Input:** random forest **M**, instance **s**, objectives **obj**
**Output:** rulelist, **clist**, comprising of local explanation of instance **s**

```
Process:
    Apply s at input interface of M
    O ← output of instance s on M
    //STEP 1: TREE SELECTION
    //Extract winning trees, WT
    WT ← ϕ
    for each tree T in M
        O_T ← output of T
        if O_T = O then
            WT ← WT + T
        end if
    end for
    //Get winning trees on the Pareto frontier, PF
    PF ← GetParetoFrontier(WT, obj)
    //STEP 2: CONDITION EXTRACTION
    //Extract conditions from PF
    clist ← ϕ
    for each tree T in PF
        for each node n in the decision path of T
            f_i ← feature at n
            op_i ← operator at n
            t_i ← threshold at n
            c ← {f_i op_i t_i}
            clist ← clist + c
        end for
    end for
    //STEP 3: RULELIST COMPRESSION
    execute intra-tree compression on clist
    execute inter-tree compression on clist
    return clist
```

2) CONDITION EXTRACTION

In this step, the rules satisfying the instance to be explained are extracted from the trees selected in step 1. Each such tree has a single path, $\{C_i \Rightarrow T\}$, that is true for the instance. The antecedent, $C_i$, consists of a conjunction of conditions on the path from the root to the respective leaf nodes. Thus, for $i$ trees on the Pareto frontier given by Eqs. 8 to 10 below, the conditions in the rule antecedents are extracted and merged into a rulelist consisting of the aggregate conjunction as follows:

$$c_{11} \cap c_{12} \cap ... \cap c_{1j} \Rightarrow T \qquad (8)$$

$$c_{21} \cap c_{22} \cap ... \cap c_{2k} \Rightarrow T \qquad (9)$$

$$c_{i1} \cap c_{i2} \cap ... \cap c_{im} \Rightarrow T \qquad (10)$$

Thus, the rulelist can be expressed as follows:

$$\text{rulelist} = c_{11} ... \cap c_{1j} \cap c_{21} ... \cap c_{2k} ... \cap c_{i1} ... \cap c_{im} \qquad (11)$$

3) RULELIST COMPRESSION

The rulelist generated in the previous step may contain repeating features. Some of these features could be sourced from different trees, while some could be sourced from the same tree, if those features are split at multiple nodes. To eliminate the repeated features, we apply two principles from previous work [34] that are applicable when the same feature and operator are present in two conditions, $c_j$ and $c_k$, with varying thresholds. The use of these principles, which we further elaborate below, ensures that the selected condition satisfies all others in the set of conditions with the same feature and operator but varying thresholds. Thus, this guarantees that the selected condition will hold across multiple conditions, which can then be subsequently eliminated from the rulelist.

a: Intra-tree condition compression (Algorithm 2)

This principle is applied to collapse rules on same features within a tree. If there are *n* number of conditions, $c_1, c_2, ... , c_n$, given by, $f\ op\ t_1, f\ op\ t_2, ... , f\ op\ t_n$, where *f* is the common feature and *op* is the common operator in these conditions and $t_1, t_2, ... , t_n$ are the respective thresholds, the most specific condition is selected. This is given by the condition with a threshold value that also satisfies the remaining $n-1$ conditions on *f* in the tree. For instance, consider three conditions on feature *f* and operator <= from a tree given by, $f <= 10$, $f <= 140$ and $f <= 250$. According to this principle the condition with the most specific threshold, i.e., $f <= 10$, is selected which also satisfies the remaining two conditions. Similarly, if there are three conditions on *f* and operator >, given by $f > 10$, $f > 140$ and $f > 250$, according to this principle $f > 250$ is selected as the condition with the most specific threshold, with a value that also satisfies the remaining two conditions.

b: Inter-tree condition compression (Algorithm 3)

This principle is applied to collapse rules on the same feature but sourced from different trees. For n number of conditions, $c_1, c_2, ... , c_n$, given by, $f\ opt_1\ t_1, f\ opt_2\ t_2, ... , f\ opt_n\ t_n$, where *f* is the common feature, *op* is the common operator and $t_1, t_2, ... , t_n$ are the respective thresholds, the most general condition is selected so that the remaining n−1 conditions continue to hold. For instance, if there are three conditions from different trees on the same feature *f* and operator <=, given by $f <= 10$, $f <= 140$ and $f <= 250$, this principle selects the most general, i.e., $f <= 250$, which also includes the range of the remaining two conditions. Similarly, if there are three conditions on feature *f* and operator >, given by $f > 10$, $f > 140$ and $f > 250$, this principle selects the most general, i.e., $f > 10$, which also includes the range of the other two conditions.

**Algorithm 2** Intra-tree condition compression

```
Input: random forest M, rulelist clist
Output: compressed rulelist clist
Process:
   for each tree T in M
      for each feature f in T
         conds ← all conditions on f
         conds_le ← filter conds on operator <=
         for each condition c in conds_le
            if threshold of c is not smallest
               clist ← clist − c
            end if
         end for
         conds_gt ← filter conds on operator >
         for each condition c in conds_gt
            if threshold of c is not largest
               clist ← clist − c
            end if
         end for
      end for
   end for
   return clist
```

**Algorithm 3** Inter-tree condition compression

```
Input: random forest M, rulelist clist
Output: compressed rulelist clist
Process:
   for each feature f in clist
      conds ← all conditions on f
      conds_le ← filter conds on operator <=
      for each condition c in conds_le
         if threshold of c is not largest
            clist ← clist − c
         end if
      end for
      conds_gt ← filter conds on operator >
      for each condition c in conds_gt
         if threshold of c is not smallest
            clist ← clist − c
         end if
      end for
   end for
   return clist
```

After applying the intra-tree (Algorithm 2) and inter-tree (Algorithm 3) compression principles, the final generated rulelist is the local explanation of the instance.

### C. EVALUATION METRICS

This subsection details the evaluation metrics for the explainability of the above proposed algorithm, the privacy of the underlying model and the overall utility of the system.

#### 1) Explainability

The explainability of the system can be described with respect to two aspects: its overall quality and its consistency in the presence of privatized training. Each of these are further detailed below:

##### a: Explanation quality

From previous works [10], [24], [35] on explanations based on explicitly scoped rules [36], we adopt the following quantitative quality metrics:

- **Precision:** This measures the fraction of instances from the dataset that satisfy the conditions and the corresponding prediction in the local explanation, given by:

$$\text{precision} = \frac{\#\text{correctly classified samples}}{\#\text{samples in a dataset}} \quad (12)$$

- **Coverage:** This measures the fraction of instances from the dataset that satisfy the conditions in the local explanation, but may or may not match the prediction. It is given by:

$$\text{coverage} = \frac{\text{\#samples that satisfy the conditions}}{\text{\#samples in a dataset}} \quad (13)$$

- **Response time per sample:** This measures the average time (in seconds) required for generation of the local explanation of an instance. For n samples, where $t_i$ is the time required for the local explanation generation of the $i^{th}$ sample, the response time per sample is given by:

$$\text{response time per sample} = \frac{1}{n}\sum_{i=i}^{n} t_i \quad (14)$$

- **Explanation length:** This measures the total number of conditions in the local explanation of an instance, given by:

$$\text{explanation length} = \text{\#conditions in the explanation} \quad (15)$$

b: Explanation consistency

To evaluate the change in explanations when privatization techniques are applied, we propose to measure the feature overlaps between the non-private and private explanations. For these overlapping features, we further propose to measure the corresponding cumulative changes in thresholds. We use the Jaccard index [37] for the former and design a new metric, average threshold change, for the latter. The combination of both metrics evaluates the drift in explainability due to DP. These metrics are defined below.

- **Jaccard index:** This measure can be used to compare sets of patterns [38]. It indicates the amount of similarity between two patterns, with its value ranging from 0 to 1. A higher value indicates higher similarity and vice versa. Given two sets, S and T, the Jaccard index is calculated as the ratio of the size of intersection to the size of union of the two sets [40], as follows:

$$J(S,T) = \frac{|S \cap T|}{|S \cup T|} \quad (16)$$

If $|S \cup T| = 0$, then $J(S,T) = 1$.

- **Average threshold change:** This measures the average normalised change in the threshold values of features contained in a rulelist with respect to a baseline. Consider a rulelist, *clist*, consisting of $n$ conditions on features, $f_1, f_2, \ldots, f_n$, with respective thresholds, $t_1, t_2, \ldots, t_n$, and a baseline rulelist, $b$, consisting of $m$ features, $f_{b1}, f_{b2}, \ldots, f_{bm}$, and respective thresholds, $t_{b1}, t_{b2}, \ldots, t_{bm}$, the threshold change of an overlapping feature, $f_i$ and $f_{bj}$, where $f_i = f_{bj}$ and matching operator is:

$$tc_i = |t_i - t_{bj}| \quad (17)$$

If $tc_{min}$ and $tc_{max}$ are the minimum and maximum threshold changes respectively for $f_i$ across all samples, the average threshold change with min-max normalisation for $k$ overlapping features is:

$$\text{ave threshold change} = \frac{1}{k}\sum_{i=1}^{k} \frac{|tc_i - tc_{min}|}{|tc_{max} - tc_{min}|} \quad (18)$$

When $tc_{min} = tc_{max}$ for $f_i$ in the above, we set its normalised value to 0. Thus, this measure gives us values in [0, 1] with lower values indicating lower drifts in the explanation thresholds and vice versa.

2) Privacy

The privacy of the DP trained models is given by the privacy budget, $\epsilon$. Due to the post-processing property of DP [16], the data independent mapping in the proposed explainability algorithm (i.e., Algorithms 1, 2 and 3) is also differentially private with the same privacy budget, $\epsilon$. The entire pipeline, including the models, predictions and explanations thus being differentially private, we test the privacy empirically by membership inference attack on the models. The effectiveness of this attack can be measured by the true positive rate (TPR) at low false positive rates (FPR) [39]. We report the log-scale ROC curves at different values of $\epsilon$ and number of trees, $\tau$.

3) Utility

The utility of the system can be determined using the accuracy of the models. The k-fold cross validation accuracy of the models with a selected value of $k$ is used for this purpose.

**D. ILLUSTRATIVE EXAMPLE**

To demonstrate our explainability method, we consider the following toy example based on a simplified version of the Credit [40] dataset:

Assume that Carol has a credit card with ABC Bank and has recently applied for an increase in her credit limit from the current limit of 10K. A credit officer evaluating her application uses the bank's tree ensemble model that predicts if she will default on the next payment, among other criteria to approve her application. For simplicity, we set the number of trees and maximum depth to 3 for this model. The model uses the following information about Carol at inference time:

- education: graduate school, marital status: single, age: 26 years, sex: female
- repayment status of last 3 months, pay 1: 0, pay 2: 0, pay 3: 0 (i.e., paid duly)
- bill amounts of last 3 months, bill amt1: 1156, bill amt2: 316, bill amt3: 316
- payment amount in last 3 months, pay amt1: 316, pay amt2: 316, pay amt3: 316

The model predicts that Carol will not default (i.e., class 0). However, to understand this decision, the officer uses the explanation module which runs the 3 steps of the algorithm.

Fig. 2 shows the trees generated in the forest with shaded nodes indicating the decision paths for the instance. In step 1, the winning trees are optimized and tree 0 is eliminated. The conditions are extracted in step 2 and after applying intra tree and inter tree compression in step 3, the selected conditions (boxed nodes) are extracted. This generates the following explanation:

sex > 1, pay 3 <= 0.0, pay 1 <= 0.0, pay 2 <= 0, bill amt2 <= 434.5

An equivalent explanation in natural language is:

Carol will not default because she is female, has not defaulted in the last 3 months and her bill amount two months ago is lower than $434.5.

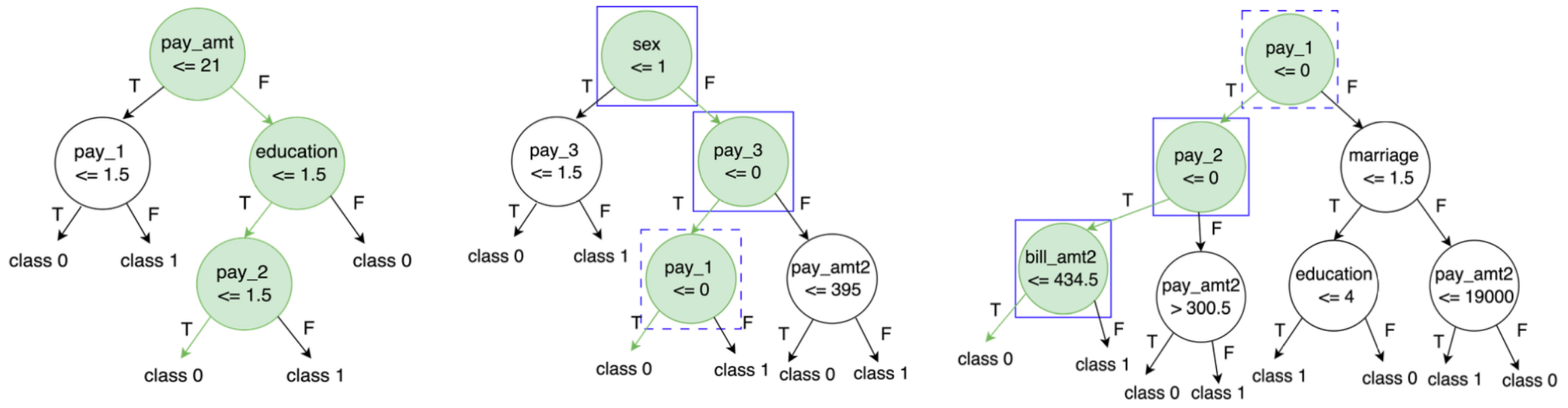

(a) Tree 0 (b) Tree 1 (c) Tree 2

FIGURE 2: Tree ensemble with 3 trees. Tree 0 is excluded from the Pareto frontier.

## IV. EXPERIMENTAL RESULTS

In this section, we elaborate the experimental settings used, including the datasets, the training details of the tree ensemble models and the specific libraries used. We also report the results observed using the evaluation metrics. All experiments were executed on a 2 GHz Quad-Core CPU with 16 GB 3733 MHz RAM.

### A. DATASETS

For the analysis in this study, datasets used in prior works on privacy preserving explanations are chosen. The properties of these datasets are summarised in Table 1 and elaborated below:

- **Adult:** This dataset consists of 48,842 records extracted from the 1994 Census database [41]. It consists of 14 features and the prediction problem is if an individual's annual income exceeds $50,000.
- **Credit:** This dataset consists of 30,000 records of default payments of individuals in Taiwan [40]. It consists of 23 features and the prediction problem is if an individual will default on payment.
- **Compas:** This dataset consists of 7,214 records of criminal defendants of Broward County, Florida, between 2013 and 2014 [42]. The dataset was collected by Propublica to examine a commercial recidivism risk assessment tool called Correctional Offender Management Profiling for Alternative Sanctions (COMPAS). We extracted a subset of 10 features for training. The prediction problem is if an individual will reoffend.
- **Student:** This dataset consists of 4,424 records with 36 features, collected to determine early risks in students' academic success and to reduce the dropout rate in higher education [43]. The prediction problem is if a student will dropout from the academic program.

TABLE 1: Dataset properties.

| Dataset | Records | Features | Records after preprocessing | Features after preprocessing |
|---|---|---|---|---|
| Adult | 48,842 | 14 | 32,561 | 44 |
| Credit | 30,000 | 23 | 30,000 | 23 |
| Compas | 7,214 | 10 | 6,907 | 15 |
| Student | 4,424 | 36 | 4,424 | 35 |

All datasets were preprocessed to remove null and missing values. Categorical columns were one-hot encoded and the respective target columns were converted to binary classification problems. The datasets were split into 70% for training and 30% for testing. The test set was further split in half, with half set aside for generating local explanations for evaluation. We randomly selected 1000 samples from this split from Adult, Credit and Compas, and 600 from the smaller Student dataset. The second half of the test set was used to simulate public data to calculate test support and confidence (see Section IIIA). We also ran an ablation study replacing this portion of the test set with synthetic data. The synthetic data was generated from the original datasets using the Gaussian Copula synthesizer from the Synthetic Data Vault library [44].

### B. MODELS

The random forest algorithm of Fletcher et al. [26] was used to train non-private and private models on each dataset. The algorithm is applicable to discrete and continuous features, as well as categorical features converted into one-hot encoding representations. We used the algorithm's implementation in IBM's diffprivlib [45] and provided the privacy budget as a parameter by varying it in the strong formal privacy guarantees (i.e., $\epsilon \leq 1$) [5], from 0.001 to 1 in multiples of 10. All models were set to a constant maximum depth of 10 and the number of trees, $\tau$, was varied from 10 to 1000, in multiples of 10. We used 10-fold cross validation on the training set and report the average accuracies of the models.

### C. MULTI-OBJECTIVE OPTIMIZATION

For determining optimized trees, we adopted the implementation of Pareto optimality from the paretoset library [33]. This library requires the objectives and the optimization direction, i.e., maximization or minimization, as parameters for determining the set of non-dominated solutions on the Pareto frontier. We configured the method on the following objectives:

- test support
- test confidence
- training accuracy
- rule length

Our proposed explainability algorithm is flexible and allows dynamic consideration of objectives, so that the above can be added or removed from the optimization process depending on the requirements of the considered problem. Additional objectives, apart from the above, can also be added to the MOO in future to adapt to the problem requirements.

## D. EVALUATIONS

### 1) EXPLAINABILITY

The evaluations in this subsection were executed on the samples set aside for explanation generation (see Section IVA). The average values on the selected metrics are reported for each combination of $\epsilon$ and $\tau$.

a: Explanation quality

On running the ablation tests on optimization using publicly available test set and synthetic data, the explanation quality metrics were found to be consistent across both these types of data. In Figs. 3 to 6, (a)-(d) show the results with optimization on the test set, while (e)-(h), show the same on synthetic data. As observed, each quality metric displayed similar trends across different values of $\epsilon$and $\tau$, irrespective of the type of data it was optimized on. Below, we further elaborate on these observations:

- **Precision:** The precision of the local explanations was found to be between 66 to 82% for non-private explanations when the number of trees was low, i.e., $\tau = 10$. This range reduced to 39 to 79% for higher number of trees, i.e., $\tau = 100, 1000$. The use of DP was found to further the adverse effect on precision values at lower privacy budgets of 0.001 and 0.01 for all considered values of $\tau$. As $\epsilon$ increased to 0.1 and 1, the precisions increased towards non-private values. Thus, both $\epsilon$ and $\tau$ had proportionate effects on precision, reducing it at low $\epsilon$ and high $\tau$ values.
- **Coverage:** The coverage of the local explanations was observed to be between 1 to 15% for all datasets except Credit, for which the coverage was higher in the range of 20 to 45%. Lower number of trees were observed to have better coverage in comparison to ensembles with larger number of trees. It was also observed to be unaffected by the privacy budgets and remained stable across different $\epsilon$ values.
- **Response time:** The average time for generation of a single local explanation was found to vary between 0.02 and 7.87 seconds for lower number of trees, i.e., $\tau = 10, 100$. For $\tau = 1000$, due to the increase in the MOO set and the subsequent increase in the conditions in the explanations, the response time was also observed to increase. This increase was higher for Adult and Credit datasets with maximum response times of 28.81 and 55.92 seconds respectively. The higher response time was due to the larger size of these datasets and the subsequent large test and synthetic datasets used for optimization, which was capped at 25% the size of the original dataset (see Section IV A). Adult and Credit were optimized on 8,140 and 7,500 records respectively, in contrast with 1,727 and 1,106 records for Compas and Student respectively. The response time can thus be shortened by reducing the records used for optimization.

- **Explanation length:** The explanation length was found to be influenced by two factors: the number of trees in the ensemble and the number of features in the dataset. It was shorter for lower τ values and increased as more optimized trees were found on the Pareto frontier. The length was also observed to be longer for datasets with more features. Thus, Compas with 15 features, had shorter explanations in contrast with Adult, Credit and Student, that had 44, 23 and 35 features respectively.

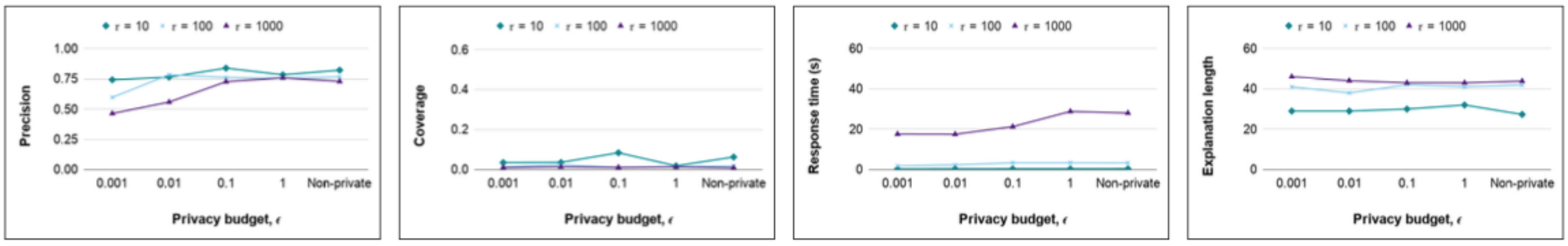

(a) Precision (b) Coverage (c) Response time (d) Explanation length

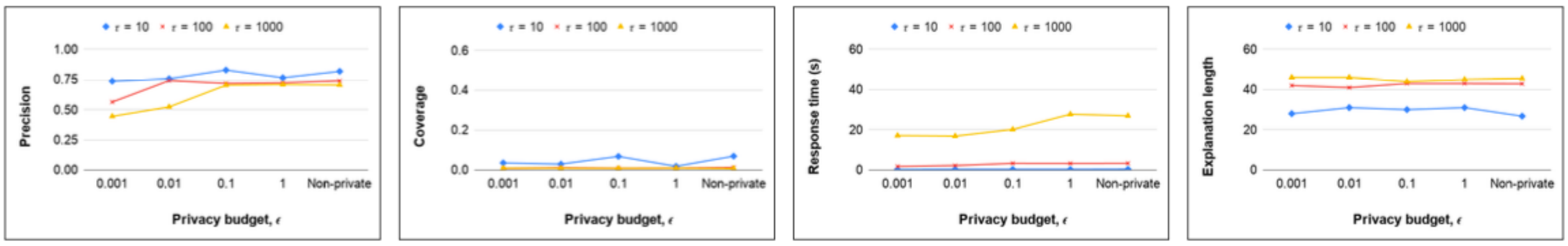

(e) Precision (f) Coverage (g) Response time (h) Explanation length

FIGURE 3: Adult dataset. (a)-(d) test set optimization, (e)-(h) synthetic data optimization.

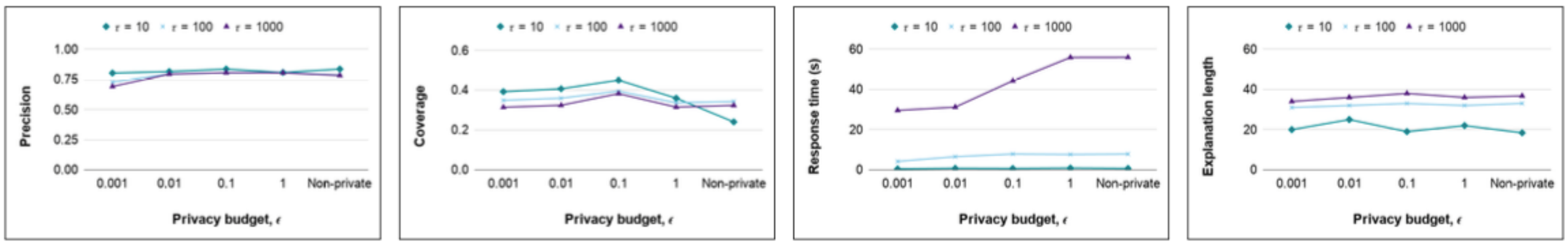

(a) Precision (b) Coverage (c) Response time (d) Explanation length

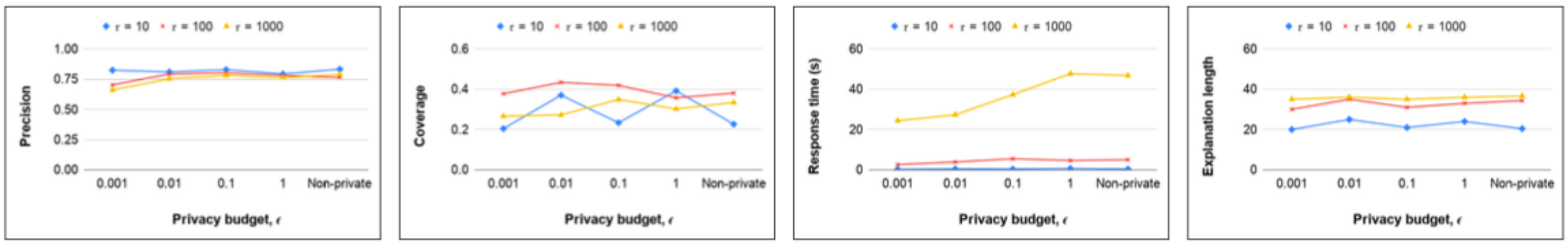

(e) Precision (f) Coverage (g) Response time (h) Explanation length

FIGURE 4: Credit dataset. (a)-(d) test set optimization, (e)-(h) synthetic data optimization.

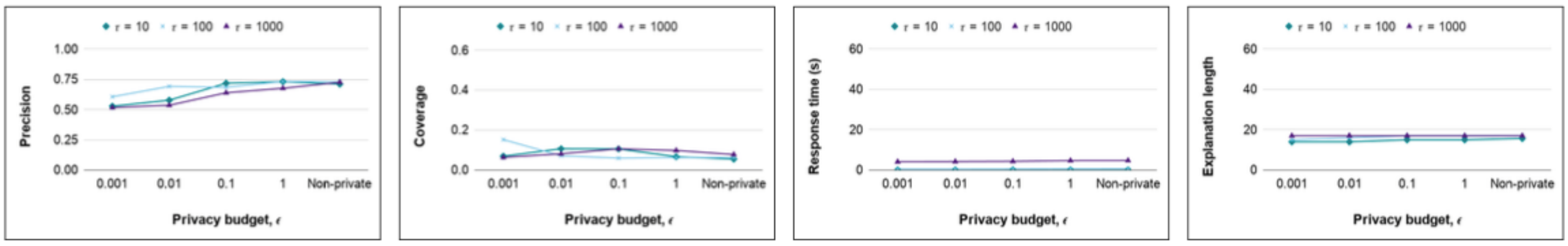

(a) Precision (b) Coverage (c) Response time (d) Explanation length

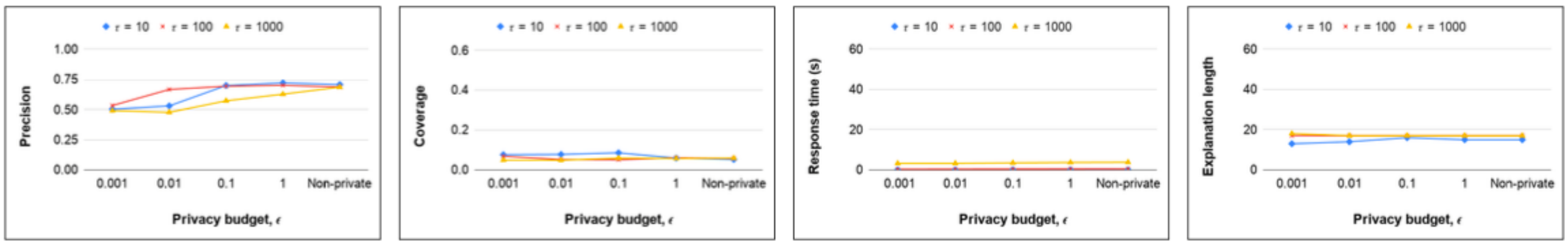

(e) Precision (f) Coverage (g) Response time (h) Explanation length

FIGURE 5: Compas dataset. (a)-(d) test set optimization, (e)-(h) synthetic data optimization.

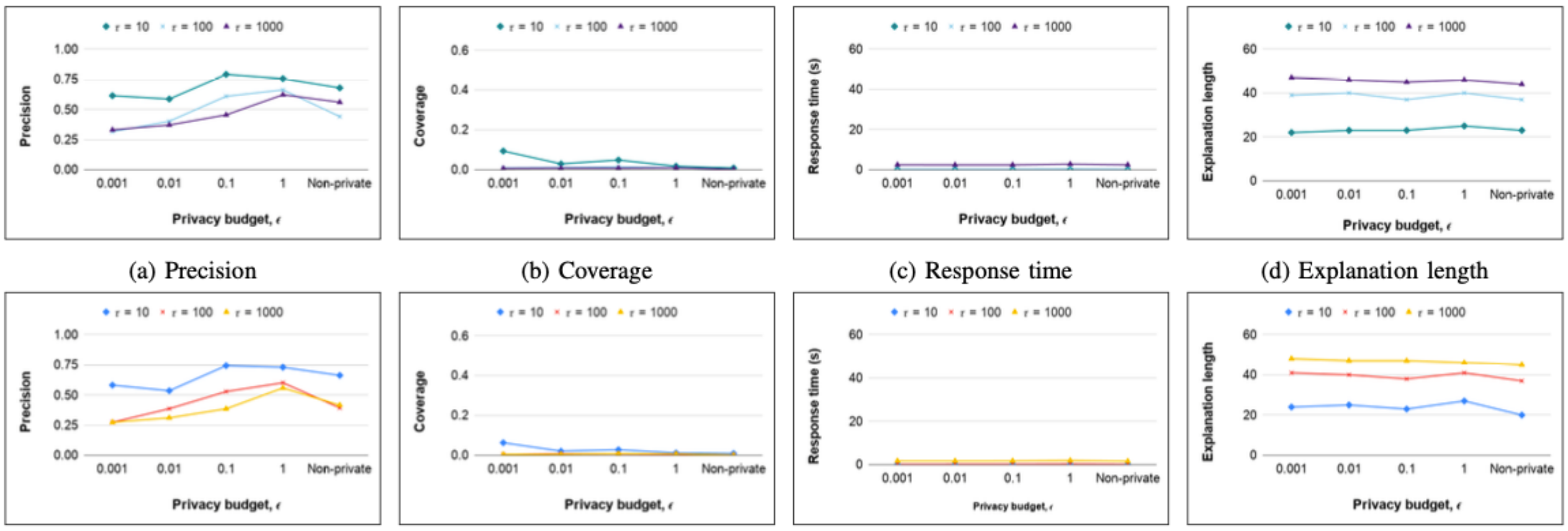

(a) Precision (b) Coverage (c) Response time (d) Explanation length

(e) Precision (f) Coverage (g) Response time (h) Explanation length

FIGURE 6: Student dataset. (a)-(d) test set optimization, (e)-(h) synthetic data optimization.

b: Explanation consistency

In this subsection, we evaluate the effect of DP training of the models on the consistency of explanations at different privacy budgets. As discussed in the previous subsection, since the explanation quality was found to be consistent across the test and synthetic datasets, we used the explanations optimized on the test set for this evaluation and report the results below.

- **Jaccard index:** As seen from Figs. 7 (a)-(d), the Jaccard index increases as the trees in the ensemble increase. It is observed to be between 0.39 to 0.87 when $\tau = 10$ and increases to the range 0.88 and 1 for $\tau = 1000$. This increase in the index can be attributed to the explanation length which correspondingly increases with higher values of $\tau$ causing the addition of more features to the explanation. This raises the number of common features between the private explanations and their non-private versions. The index, however, is found to be unaffected by the privacy budgets and stays consistent across different $\epsilon$ values.
- **Average threshold change:** As seen from Figs. 7 (e)-(h), the average change in the thresholds of the conditions in the explanations is between 0.19 and 0.49. Notably, it is observed to decrease as $\tau$ increases, with lower number of trees displaying the highest values. Similar to the Jaccard index, this measure is also consistent across different privacy budgets. Since this measure indicates the changes in the thresholds of private explanations with respect to their non-private counterparts, we note non-negligible changes to the thresholds when DP is introduced in the explanation pipeline.

2) PRIVACY EVALUATION

The privacy evaluation of the explanation pipeline was empirically observed by executing the Likelihood Ratio Attack (LiRA) [39] directly on the models. We used the implementation of this attack in the SACRO-ML package [46],[47]. Each non-private model and its private counterpart, trained with $\epsilon$ in [0.001, 0.01, 0.1,1] and $\tau$ in [10, 100, 1000], and its non-private counterpart, was attacked using 100 trained shadow models. Fig. 9 in the appendix, shows the ROC curves for TPR at low FPR generated by the toolkit for the non-private models and their private counterparts at the end-points of the considered privacy range, i.e., $\epsilon = 0.001$ and 1. The diagonal line in the ROC curve indicates random guess. Parts of the curve above the diagonal line indicate successful attacks, while those below indicate successful mitigation. From the ROC curves for different $\epsilon$ and $\tau$ values, we observed that the DP trained models were successful in holding the membership inference success rates at random guess. In many cases, where the success rate was above random guess, such as the non-private model for Student in Fig. 9 (l), DP was observed to successfully mitigate the attack. Notably, the non-private models for the remaining 3 datasets were found to be resilient to membership inference attacks, which can be attributed to the random selection of features for node splitting during the tree building process in the selected random forest algorithm.

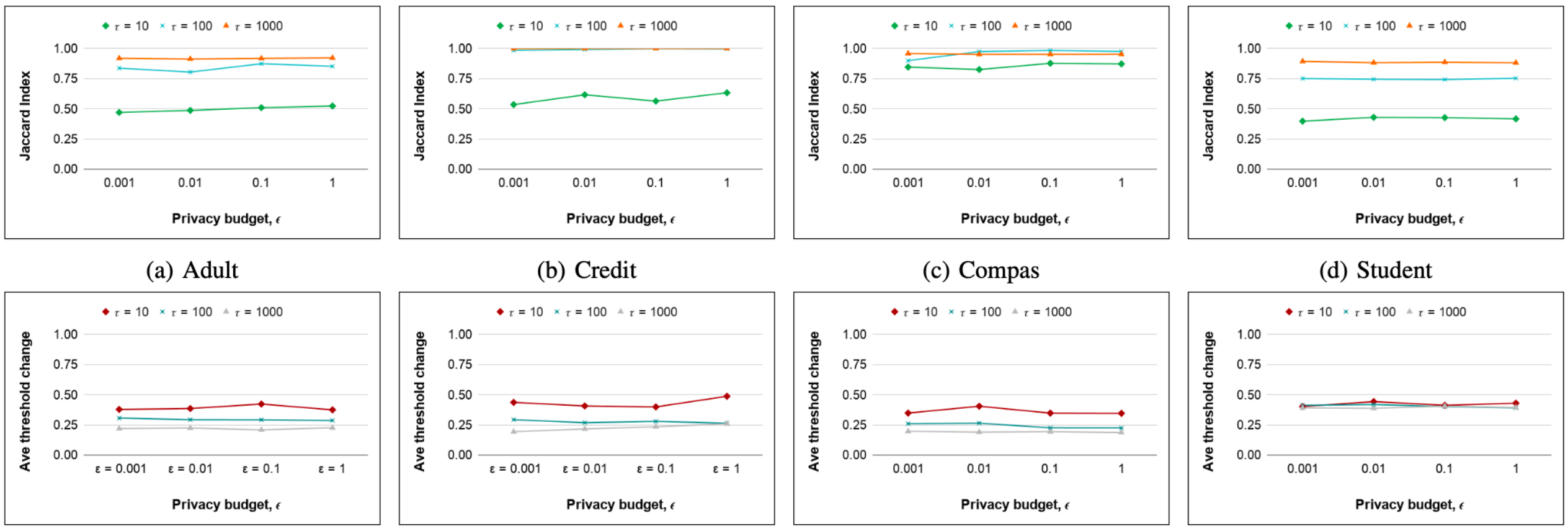

(a) Adult (b) Credit (c) Compas (d) Student

(e) Adult (f) Credit (g) Compas (h) Student

FIGURE 7: Effect of $\epsilon$ and $\tau$ on (a)-(d) Jaccard index, (e)-(h) Average threshold change.

### 3) UTILITY EVALUATION

The accuracies of the models trained non-privately is summarised in Table 2. As observed, increasing the number of trees had negligible effect with accuracies remaining stable across different values of $\tau$. The use of DP training, however, had non-negligible effect on the accuracies as seen from Fig 8. The accuracies were found to be highest for the non-private models for all datasets. These values decreased at very low budgets below 0.01, and stabilized at budgets $\geq 1$ approaching the accuracies of the non-private models.

TABLE 2: Test accuracies (%) of non-private models.

| Dataset | $\tau = 10$ | $\tau = 100$ | $\tau = 1000$ |
|---|---|---|---|
| Adult | 76.10 | 75.36 | 75.26 |
| Credit | 78.24 | 77.77 | 77.73 |
| Compas | 70.02 | 71.68 | 70.89 |
| Student | 86.08 | 86.98 | 87.15 |

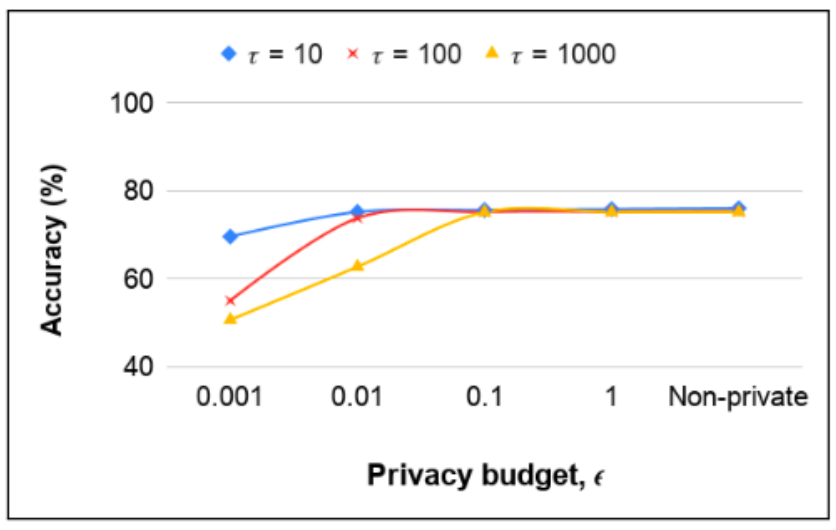

(a) Adult

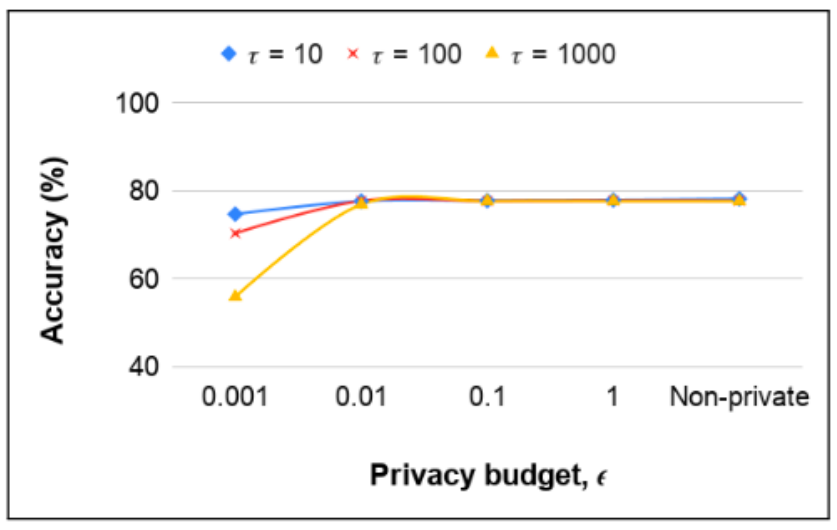

(b) Credit

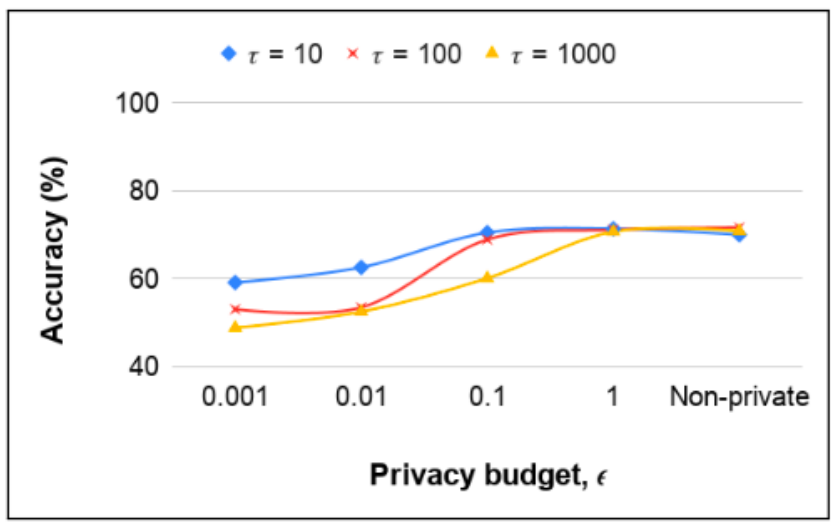

(c) Compas

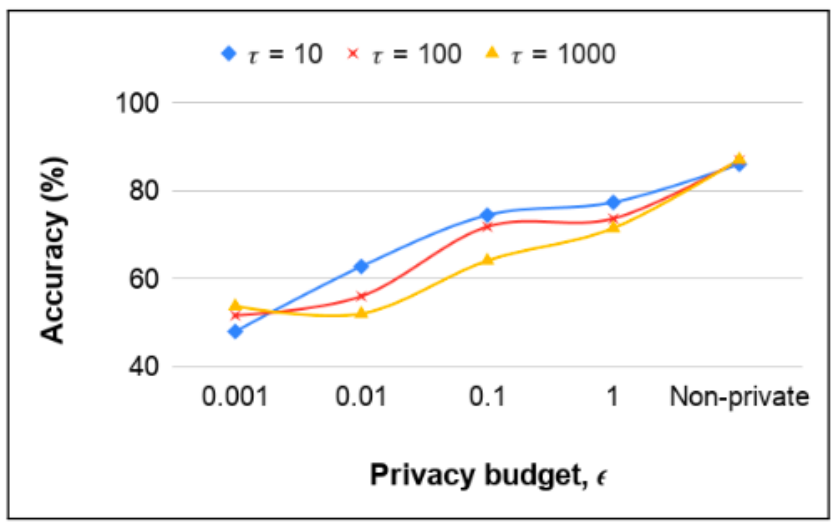

(d) Student

FIGURE 8: Effect of $\epsilon$ on random forest accuracy.

## V. DISCUSSION

Our proposed method of building explainability for an instance, uses the intrinsic components of a model optimized over a combination of objectives on the training and test sets. To study the effect of using publicly available test sets on the optimization process, we conducted an ablation study with synthetic data and found consistent explanation quality across the usage of both types of datasets. Since the usage of synthetic data has no additional benefits on the explanation quality, we propose the use of test sets for optimization to reduce the additional computational cost of generating synthetic data.

A notable feature of our explainability algorithm is that since the individual interpretable components are drawn from within the architecture, the explanations are reflective of the model and are faithful by design. There is no approximation in the optimization process and the trees selected through Pareto optimality process are non-dominated on the selected objectives. Thus, none of the optimized trees can be improved further on the selected objectives without deteriorating other objectives. The algorithm is flexible in its inclusion and exclusion of these objectives, which can be tailored according to the requirements of the domain and the architecture of the underlying models. This approach differs from current explainability strategies that use feature importances [48], examples [49], concepts [50] and knowledge-based approaches [51] to craft explanations. Our approach is more aligned towards mechanistic interpretability [52] approaches that seek to reverse engineer models from individual interpretable components in neural networks.

The local explanations for specific instances determined using our method have coverage of up to 45% on the training set and precisions up to 82%, demonstrating their generalisation ability. The explanations have a short turnaround time making them feasible for real world applications, where explanations are required quickly for practical deployments. Since the length of explanations is associated with complexity [10] and longer explanations tend to deteriorate the understanding to end-users, our method for tree ensembles can be used with smaller number of trees to generate shorter explanations without deteriorating the accuracy of the model. This choice of the number of trees can be taken into consideration during the model design phase of the AI lifecycle. Thus, in applications where concise explanations are needed for points of interest, ensembles with fewer trees can be deployed.

Explanations are found to leak the privacy of individuals and DP is an attractive privacy enhancing technology due to its mathematical privacy guarantee. However, it is found to have non-negligible effects on model accuracies and drifts in explanations. For our proposed algorithm, in addition to observing the effect of different privacy budgets on model accuracies, we proposed metrics for quantifying the consistency in explanations between non-private and private models. The Jaccard index measured the feature overlap between the non-private and privatized versions, and was observed to increase as the number of trees in the ensemble increased. For these overlapping features given by the Jaccard index, we also measured the average normalised change in the thresholds, which was found to be non-negligible. Both measures were found to be independent of the privacy budget and dependent on the number of trees. Higher number of trees were found to have higher Jaccard index and lower average threshold change, and this can be a design consideration when selecting the number of trees in the model. However, this requirement can conflict with explanation length which also increases as the trees are increased. To further investigate this interplay of conflicting requirements, a human-centred evaluation [56] of these explanations can shed further insights in determining the right balance between privacy and users' understanding.

Though, currently we have demonstrated our algorithm on tree ensembles based on majority voting, in future we will work on extending its applicability to other model architectures, such as deep networks, by isolating the interpretable components from their architectures and designing the model-specific optimization objectives. Our proposed metrics for determining the consistency of explanations in the presence of privacy enhancing technologies, is applicable to other local and global explanations that derive rulesets comprising of conjunction of conditions to describe model behaviour and outcomes. Explanation methods using such explicitly scoped rules [36], such as anchors [54], LORE [35] and LionForests [24], fall under this category. In future work, we will run a comparative analysis of our method on quality and consistency with these explanation methods. We will also explore the integration of natural language explanations and their impact on the understanding of explanation users.

## VI. RELATED WORKS

The use of black-box algorithms for automated AI decision-making in high-stakes domains is a detriment to the trustworthiness of these systems, since transparency and explainability are the foundational principles for responsible and ethical systems. To overcome this challenge, researchers have proposed different XAI strategies for explaining the behaviour of black-box models and their outcomes. Mechanistic interpretability is one such approach for understanding the functioning of neural networks [27], [55], based on isolating activations, features and circuits that influence the outcomes. However, this concept has been primarily considered in connectionist architectures and its applicability to other model architectures has been unexplored.

Specifically to tree ensembles, which are also black-boxes due to the use of multiple decision trees, explainability methods such as rule extraction [24], [30], [56] and tree approximations [57]–[59] have been studied. Rule extraction approaches are useful in mining the discovered knowledge, in the form of rulesets which provide insights into the model's outcomes. In contrast, tree approximation approaches aim at condensing the ensemble into a single decision tree to model the relationship between features and predictions, while retaining the original predictive accuracy. In addition to these two mainly explored techniques, another approach involves the building of semantically equivalent algebraic decision diagrams [60] to form local and global explanations. Our work differs from previous work in its application of mechanistic interpretability concepts in tree ensembles for the formulation of local explanations to provide the rationale of automated decision-making to the recipients of AI systems. Our approach consolidates optimized nodes using a combination of objectives and is faithful by design due to the use of the internal components of the model.

Recent research has proposed the integration of privacy enhancing technologies in XAI methods [61] to mitigate the leakage of the privacy of individuals through explanations. DP is one such widely adopted technique for interpretable, feature-based and example-based XAI methods [6]–[8], [62], [63]. However, in addition to affecting the utility of models, it is also found to adversely affect the explanation quality [62]. Hence this work differs from previous research in the careful consideration of DP on the triad of explainability, privacy and utility, which we quantify and evaluate with our proposed metrics.

## VII. CONCLUSION

In this study, we proposed a local explainability method using multi-objective optimization of model-specific objectives to determine intrinsic interpretable components of a model. We demonstrated the method on tree ensembles that use majority voting. The model-specific optimization objectives designed for the algorithm leveraged a combination of decision tree and rule properties to enable the screening of skillful trees and nodes. We also executed an ablation study using publicly available test set and synthetic data for optimization and found the explanation quality consistent across both these types of data. The algorithm was applied on random forest models trained with strong formal privacy guarantees and explainability was measured in terms of its quality and consistency with non-private explanations. Though DP was effective at mitigating membership inference attacks and holding the success rates at random guess, we found non-negligible change in the explanations due to the privatized training and drop in the model accuracies at strong privacy guarantees. The drift in the explanations was influenced by the number of trees in the ensemble, rather than the privacy budget, hence this can be a design consideration during the model creation stage.

Though in the current work we have demonstrated the applicability of mechanistic interpretability in tree ensembles, in future we will extend the method to other model architectures. We aim to integrate natural language into the explanations and conduct a human-centred explainability evaluation to determine the effect of privacy preserved explanations on end-users. In future work, we will also do a comparative study of our approach with other explainability methods, that use similarly explicitly scoped rules, on the properties of explainability, privacy and utility.

## APPENDIX

See Fig. 9.

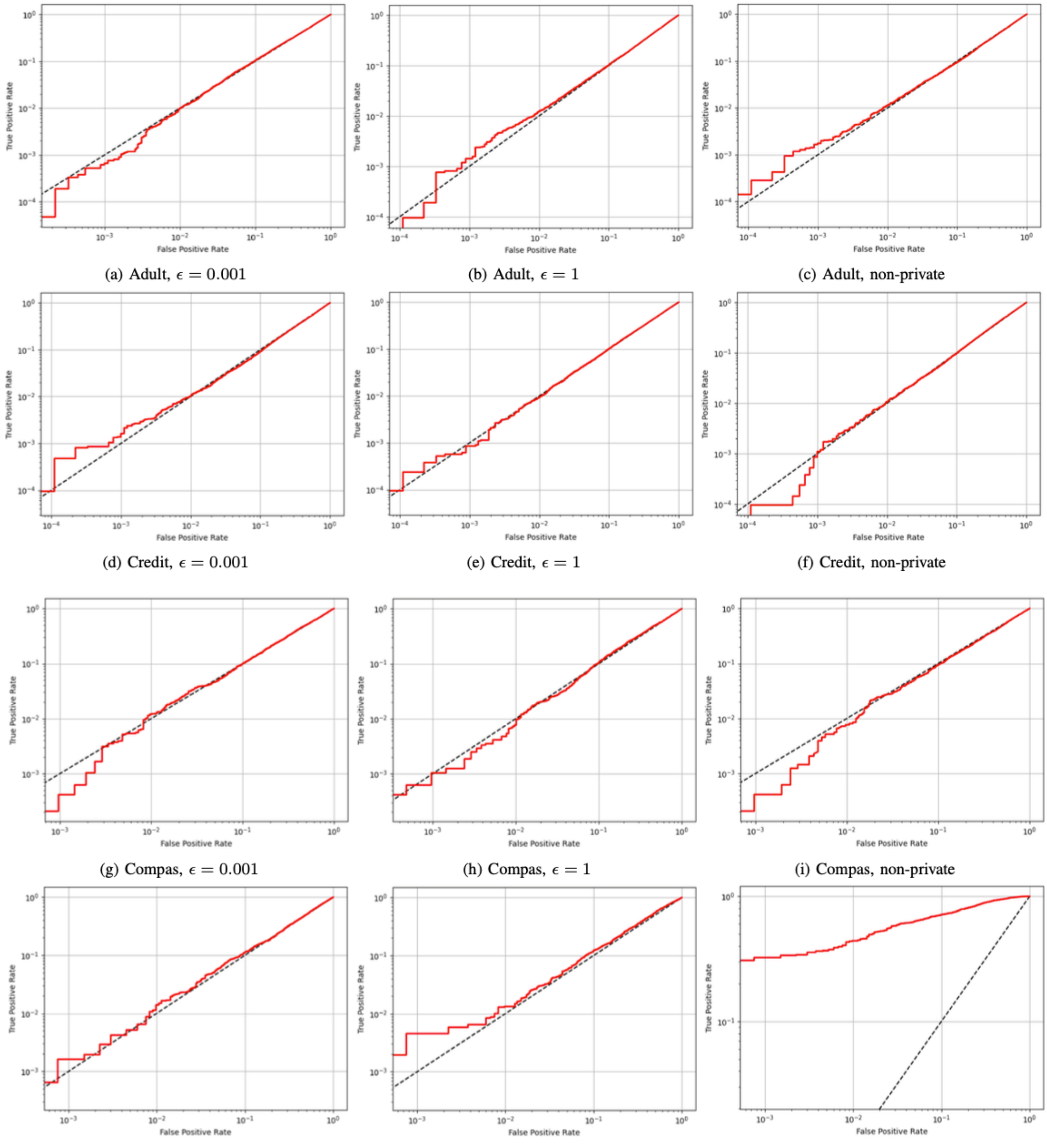

(a) Adult, $\epsilon = 0.001$
(b) Adult, $\epsilon = 1$
(c) Adult, non-private
(d) Credit, $\epsilon = 0.001$
(e) Credit, $\epsilon = 1$
(f) Credit, non-private
(g) Compas, $\epsilon = 0.001$
(h) Compas, $\epsilon = 1$
(i) Compas, non-private
(j) Student, $\epsilon = 0.001$
(k) Student, $\epsilon = 1$
(l) Student, non-private

FIGURE 9: ROC curves (TPR at low FPR) for membership inference attacks at different privacy budgets ($\tau = 1000$).